\documentclass[10pt,twocolumn,letterpaper]{article}

\usepackage[pagenumbers]{cvpr} % To force page numbers, e.g. for an arXiv version
\usepackage{multirow}
\usepackage{algorithm}
\usepackage{algpseudocode}

\definecolor{cvprblue}{rgb}{0.21,0.49,0.74}
\usepackage[pagebackref,breaklinks,colorlinks,allcolors=cvprblue]{hyperref}

\def\paperID{39675} % *** Enter the Paper ID here
\def\confName{CVPR}
\def\confYear{2026}

\title{NeuroFlow: Toward Unified Visual Encoding and Decoding from Neural Activity
}

\author{
Weijian Mai$^{1,2}$ \quad
Mu Nan$^{2,3}$ \quad
Yu Zhu$^{1}$ \quad
Jiahang Cao$^{1,2}$ \quad
Rui Zhang$^{2}$ \quad
Yuqin Dai$^{4}$ \\
Chunfeng Song$^{1\dagger}$ \quad
Andrew F. Luo$^{2\dagger}$ \quad
Jiamin Wu$^{1,5\dagger}$ \\[0.3em]
$^1$Shanghai Artificial Intelligence Laboratory \quad
$^2$University of Hong Kong \\
$^3$Shenzhen Loop Area Institute \quad
$^4$Tsinghua University \quad
$^5$Chinese University of Hong Kong \\[0.3em]
\textit{Project Page: \href{https://michaelmaiii.github.io/NeuroFlow-S}{https://michaelmaiii.github.io/NeuroFlow-S}}
}

\begin{document}
\maketitle

\begin{abstract}
% Visual encoding and decoding models serve as a computational interface linking neural activity to human visual perception.
\noindent Visual encoding and decoding models act as fundamental tools for understanding the neural mechanisms underlying human visual perception. 
Typically, visual encoding models that predict brain activity from stimuli and decoding models that reproduce stimuli from brain activity are treated as distinct tasks, requiring separate models and training procedures. 
This separation is inefficient and fails to model the consistency between encoding and decoding processes.
To address this limitation, we propose \textbf{\textit{NeuroFlow}}, the first unified framework that jointly models visual encoding and decoding from neural activity within a single flow model. 
NeuroFlow introduces two key components: (i) \textbf{NeuroVAE} is designed as a variational backbone to model neural variability and establish a compact, semantically structured latent space for bidirectional modeling across visual and neural modalities.
(ii) \textbf{Cross-modal Flow Matching (XFM)} bypasses the typical paradigm of noise-to-data diffusion guided by a specific modality condition, instead learning a reversibly consistent flow model between visual and neural latent distributions. For the first time,\textbf{ visual encoding and decoding are reformulated as a time-dependent, reversible process within a shared latent space} for unified modeling.
Empirical results demonstrate that NeuroFlow achieves superior overall performance in visual encoding and decoding tasks with higher computational efficiency compared to any isolated methods.
We further analyze principal factors that steer the model toward encoding–decoding consistency and demonstrate through brain functional analyses that NeuroFlow captures consistent activation patterns underlying neural variability.
NeuroFlow marks a major step toward unified visual encoding and decoding from neural activity, providing mechanistic insights that inform future bidirectional visual brain–computer interfaces.
% Code and project page are publicly available at \href{https://michaelmaiii.github.io/NeuroFlow-S}{NeuroFlow}.

\end{abstract}
\vspace{-4mm}
\section{Introduction}
\label{introduction}
Understanding the neural mechanisms of visual encoding and decoding is fundamental to neuroscience~\cite{word-select, speech-select, nsd-encoding, multimodel-encoding} and essential to advancing brain-computer interfaces~\cite{fernandes2024considerations}.
Visual encoding aims to transform external stimuli into neural activity, whereas visual decoding reverses this transformation to recover perceptual content from neural responses.
Functional magnetic resonance imaging (fMRI), a non-invasive neuroimaging technique, has emerged as a promising modality for visual encoding and decoding from the human brain, as it measures blood-oxygen-level-dependent signals with high spatial resolution that serve as indirect proxies for neural activity~\cite{naselaris2011encoding}.
By translating visual inputs into fMRI-measured neural patterns, visual encoding models not only advance our understanding of human visual perception, but also lay the biological groundwork for research in visual decoding.

%%现有研究做法：i) Encoding, ii) Decoding, iii) Encoding & Decoding
Recent advances in this domain predominantly focus on either encoding or decoding in isolation: (i) \textbf{Encoding models}~\cite{luo2024brain,mai2025synbrain,bao2025mindsimulator} predict neural responses given visual input, thereby characterizing stimulus-driven brain representations; (ii) \textbf{Decoding models} reproduce visual stimuli based on brain activity, providing a window into perceptual content~\cite{mai2024brain}. 
Several models attempt to connect both directions, but still struggle with learning \textit{two} \textit{independent networks} to bridge the visual and neural domains between pixel-voxel spaces~\cite{beliy2019voxels,gaziv2022self} or latent spaces~\cite{qian2024lea}.
These challenges motivate a key research question: \textbf{\textit{How can we unify visual encoding and decoding within a single model?}}
%%What property should a unified model possess?
A unified model should satisfy two essential properties: \textbf{(i) Shared latent space:} encoding and decoding processes should be optimized in a common latent space that supports mutual interactions; \textbf{(ii) Encoding-decoding consistency:} encoding and decoding serve as complementary processes that should ensure synthetic neural signals can be transformed back into coherent images in the reverse direction.

\begin{figure}[!t]
\begin{center}
\centerline{\includegraphics[width=0.47\textwidth]{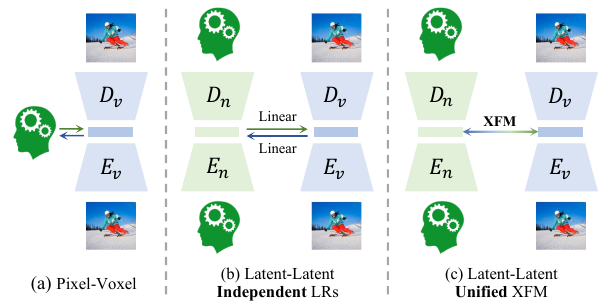}}
\vspace{-2mm}
\caption{Visual encoding-decoding frameworks.
}
\label{fig:2}
\end{center}
\vskip -0.5in
\end{figure}

To address these challenges, we propose \textbf{NeuroFlow}, the first unified framework that jointly models visual encoding and decoding from fMRI activity within a single flow model.
NeuroFlow comprises two key components designed to satisfy essential properties of unified modeling. 
\textbf{(i) NeuroVAE}, a variational backbone designed to achieve a compact and structured latent space for bidirectional modeling. 
NeuroVAE introduces probabilistic learning to model neural variability and constrains the latent space with visual semantics for cross-modal alignment.
This framework projects fMRI signals into a semantically organized latent space, from which neural signals can be reproduced with semantic coherence, rather than overfitting to voxel-level noise.
Hence, NeuroVAE \textit{lays the foundation for encoding-decoding consistency} by supporting visual-conditional fMRI synthesis that preserves semantic coherence.
\textbf{(ii) }\textbf{Cross-model Flow Matching (XFM)} is designed to unify encoding and decoding processes by learning a seamless and consistent flow between empirical visual and neural latent distributions. 
Here, encoding and decoding are reformulated as a \textit{time-dependent, reversible} process within \textit{a shared latent space}, where \textit{encoding-decoding consistency} is rigorously enforced by flow matching principles, and reversing the temporal direction naturally transitions between the two processes.

\begin{figure}[!t]
\begin{center}
\centerline{\includegraphics[width=0.47\textwidth]{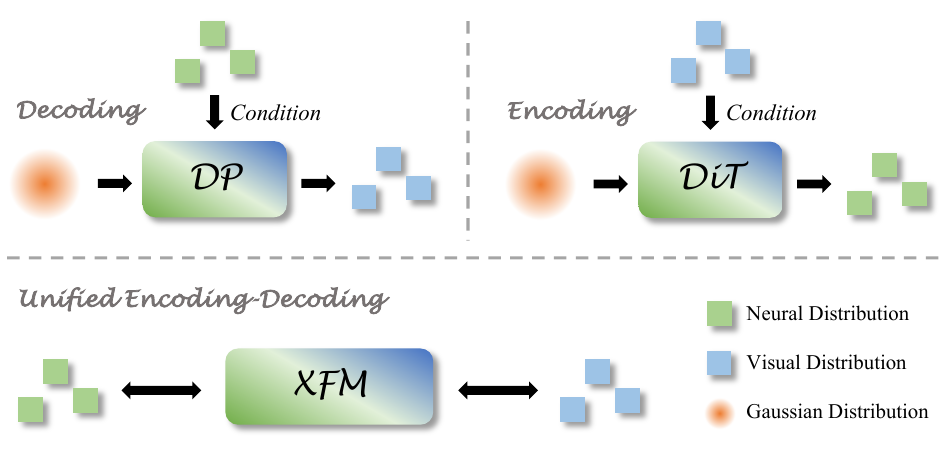}}
\vspace{-2mm}
\caption{Cross-modal alignment pipelines.
}
\label{fig:3}
\end{center}
\vskip -0.5in
\end{figure}

Through this architecture, NeuroFlow provides a unified modeling paradigm that ensures encoding-decoding consistency within a single model, as illustrated in Fig.~\ref{fig:teaser}. 
Specifically, its advantages are reflected in two aspects: \textbf{(i) Superior overall performance with strong encoding-decoding consistency}: As a unified model, NeuroFlow achieves superior or comparable performance compared with decoding methods and surpasses all encoding methods significantly, demonstrating strong consistency between encoding and decoding.
\textbf{(ii) Superior parameter efficiency: }NeuroFlow achieves strong parameter efficiency, using only 25\% of the parameters of the decoding model MindEye2, while matching or even exceeding its performance across several metrics.
Our main contributions are as follows:

\begin{itemize}
    \item We propose \textbf{NeuroFlow}, the first unified framework that jointly models visual encoding and decoding with strong semantic consistency within a single architecture. NeuroFlow achieves superior overall performance in encoding and decoding tasks with higher parameter efficiency.
    \item NeuroFlow establishes \textbf{\textit{a compact, probabilistic, and semantically organized latent space}} for bidirectional modeling across visual and neural modalities, from which neural signals can be synthesized with semantic coherence, rather than overfitting to voxel-level noise.  
    \item  NeuroFlow bypasses the conditional noise-to-data formulation and establishes \textbf{\textit{a reversibly consistent flow between visual and neural latent distributions}}. For the first time, we reformulate encoding and decoding as a time-dependent process within a shared latent space, which is reversible by simply changing the temporal direction.
    \item NeuroFlow identifies key factors essential for unified modeling and, through brain functional analysis, reveals that synthetic fMRI signals preserve \textbf{\textit{interpretable cortical patterns}} underlying biological neural variability.
\end{itemize}

\vspace{-1mm}
\section{Related Work}
\label{sec:related_work}
\vspace{-1mm}

\subsection{Visual Encoding and Decoding}
Computational modeling of neural activity typically follows two complementary paradigms: \emph{encoding} models that map from stimuli to neural responses, and \emph{decoding} models that infer stimuli from neural data~\citep{naselaris2011encoding,kamitani2005decoding,norman2006beyond,han2019variational,seeliger2018generative,shen2019deep,ren2021reconstructing,adeli2023predicting,gifford2024opportunities,dai2025mindaligner}. 
Recent advances in machine learning have accelerated progress in both directions. 

\vspace{-5mm}
\paragraph{Visual Encoding.} The prevailing recipe for encoding models couples a pretrained visual feature extractor with linear voxel-wise weights~\citep{dumoulin2008population,gucclu2015deep,klindt2017neural,eickenberg2017seeing,wen2018neural,gaziv2022self}. 
Beyond prediction, encoding models have facilitated investigations into the coding properties in higher-order visual areas~\citep{khosla2022high,khosla2022characterizing,yang2024brain,yang2024alignedcut,luo2024brain,sarch2023brain,lappe2024parallel,yu2025meta}. 
More recently, several works reformulate encoding as a \emph{generative} problem that synthesizes fMRI responses conditioned on visual input, leveraging transformer architectures~\citep{mai2025synbrain} and diffusion transformers~\citep{bao2025mindsimulator}.

\vspace{-5mm}
\paragraph{Visual Decoding.}
Decoding models provide a window into perceptual content by reproducing visual stimuli from evoked neural activity measured with fMRI, electroencephalography (EEG), and magnetoencephalography (MEG) signals~\citep{takagi2023high,chen2023seeing,lu2023minddiffuser,ozcelik2023natural,doerig2022semantic,ferrante2023brain,liu2023brainclip,mai2023unibrain,scotti2024mindeye2,benchetrit2023brain,li2024visual,guo2025neuro,xia2024umbrae,wang2024mindbridge,dream,shen2024neuro,mindvideo,lite-mind,zhou2025csbrain,lu2025unimind}.
Recent approaches typically map neural signals into pretrained vision-language embedding spaces (e.g., CLIP~\cite{clip}) and decode them back into images using pretrained generative models~\cite{rombach2022high,xu2023versatile,controlnet,ye2023ip}. 

\vspace{-5mm}
\paragraph{Visual Encoding \& Decoding.}
%结合Fig.2
Prior research has predominantly modeled visual encoding and decoding as independent problems. Recent approaches have begun to link the two, yet they still depend on distinct networks and lack a shared latent space~\cite{beliy2019voxels,gaziv2022self,qian2024lea} as illustrated in Fig.~\ref{fig:2}.
Specifically, early attempts focused on fine-grained mapping between \textbf{pixel and voxel spaces}, which often produced blurry reconstructions lacking semantic coherence~\cite{beliy2019voxels,gaziv2022self}.
To address this limitation, later studies sought to bridge the neural and visual \textbf{latent spaces} to enrich semantic information, but still rely on two independent linear regressions for the encoding and decoding directions without shared representations~\cite{qian2024lea}.
As a result, none of these methods has achieved a unified framework capable of coherent encoding and decoding transformation within a shared latent space.
% Consequently, existing methods still fall short of providing a unified framework that performs encoding and decoding in a coherent and reversible manner.

\vspace{-1mm}
\subsection{Cross-Modal Alignment}
A central challenge in visual encoding or decoding lies in \textit{cross-modal alignment} that aims to establish a precise mapping between neural and visual distributions. 
Early methods relied on simple linear regressions to approximate the unidirectional relationship, which limited their ability to capture complex semantic correspondences~\cite{takagi2023high,mai2023unibrain,ozcelik2023natural}. 
Recent approaches introduced nonlinear mappings using Diffusion Transformer (DiT)~\cite{peebles2023scalable} or Diffusion Prior (DP)~\cite{dalle2} under generative objectives~\cite{bao2025mindsimulator,scotti2024mindeye2}, operating by conditioning Gaussian noise on one modality (e.g., neural or visual latent distribution) and iteratively guiding it toward the target distribution, as illustrated in Fig.~\ref{fig:3}. 
However, such conditional noise-to-data pipelines still treat encoding and decoding as separate processes. 
In contrast, our proposed XFM establishes continuous and reversible flows directly between the neural and visual distributions, achieving a unified framework for encoding and decoding.

% The proposed \textbf{\textit{SynBrain}} framework addresses these limitations by employing probabilistic learning to model neural variability and introducing a one-step point-to-distribution mapping mechanism, allowing stable and semantically aligned fMRI synthesis in a single forward pass.
\vspace{-1mm}
\section{Methodology}
\begin{figure*}[!t]
\begin{center}
\centerline{\includegraphics[width=\textwidth]{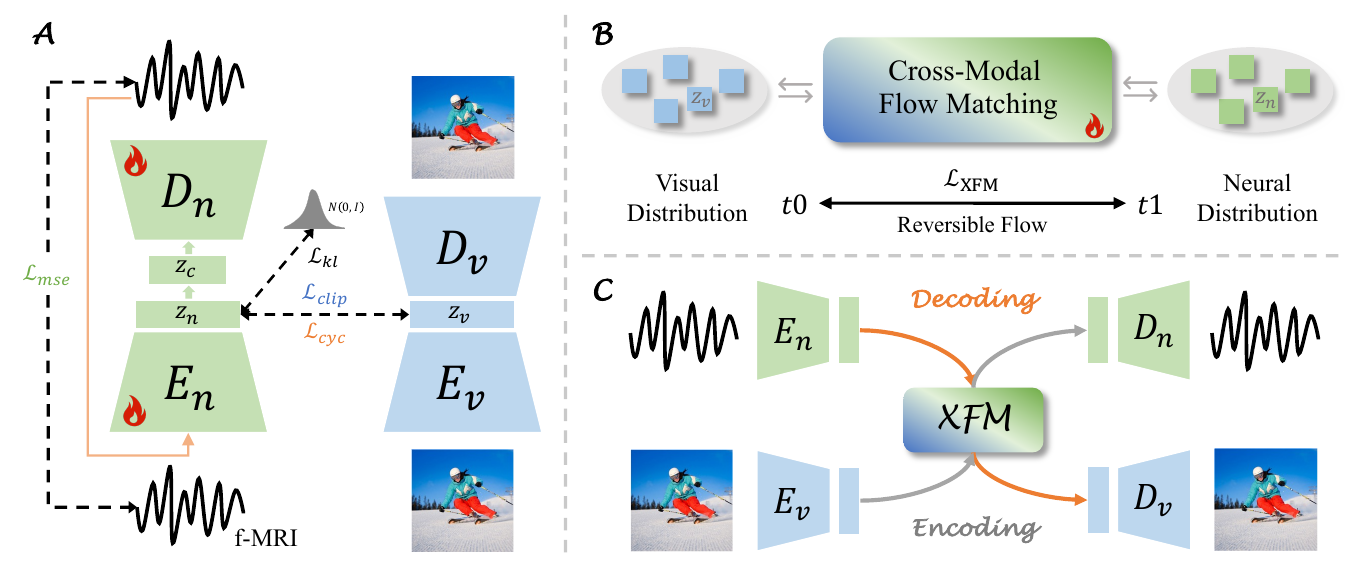}}
\vspace{-3mm}
\caption{\textbf{Architecture overview.} \textbf{Stage-1 (A):} NeuroVAE introduces probabilistic learning to model neural variability and constrains the latent space with visual semantics for consistent image-to-fMRI synthesis. 
\textbf{Stage-2 (B):} XFM unifies encoding and decoding processes by learning a time-dependent, reversible flow between empirical visual and neural latent distributions.
\textbf{Stage-3 (C):} Encoding and decoding are performed within a single model at inference, where reversing the temporal direction naturally transitions between the two processes.
}
\label{fig:overview}
\end{center}
\vskip -0.4in
\end{figure*}

\vspace{-1mm}
Our goal is to develop a unified framework for visual encoding and decoding within a single model that preserves semantic consistency between the two complementary processes.
To this end, we propose NeuroFlow, which integrates a pretrained visual backbone, a neural backbone (i.e., NeuroVAE), and a cross-model flow matching (XFM) module to bridge the gap between visual and neural latent distributions. The overall framework is depicted in Fig.~\ref{fig:overview}, and we will detail the principles and architectures in this section.

\subsection{Visual Backbone: CLIP \& UnCLIP}

The visual backbone is built upon an encoder-decoder architecture that projects visual inputs into a latent space and decodes them back into the pixel space. We instantiate this architecture using CLIP~\cite{clip} and UnCLIP\cite{dalle2,scotti2024mindeye2} models. 
The \textbf{CLIP} embedding space is trained via vision-language contrastive learning to form a high-dimensional, structured semantic space, which captures the semantic relationship between semantic concepts.
% Rather than encoding individual concepts as isolated points, it captures a \textbf{sparse yet coherent distribution} of visual semantics, preserving meaningful relationships between concepts. 
% This structured geometry enables smooth transitions between semantically related images and provides a stable coordinate system for aligning neural representations.
Specifically, a pretrained CLIP-Image encoder maps an input image $x_{img} \in \mathbb{R}^{h \times w \times 3}$ to a visual latent representation $z_v = E_v(x_{img}) \in \mathbb{R}^{m \times d}$, where $m$ is the number of visual tokens and $d$ is the feature dimension. 
A pretrained \textbf{UnCLIP} model serves as the generative counterpart to synthesize images from visual latents $\hat{x}_{img} = D_v(z_{v})$. This visual backbone is kept \textbf{frozen} in this work, offering a \textbf{semantically grounded interface} for bridging visual and neural modalities.

\subsection{Neural Backbone: NeuroVAE}

NeuroVAE is a variational backbone designed to model the neural variability and to achieve a compact, structured latent space for bidirectional modeling between fMRI and stimuli. 
As illustrated in Fig.~\ref{fig:overview}-A, NeuroVAE introduces probabilistic characteristics to model the \textbf{\textit{one-to-many}} relationships between visual stimuli  and fMRI responses \cite{mai2025synbrain} , and constrains the latent space with visual semantics for structured distributions and semantically coherent reconstructions.

Given an fMRI input $x_{\mathrm{fMRI}} \in \mathbb{R}^{1 \times n}$, the neural encoder $E_n$ estimates a posterior distribution $q (z_n \in \mathbb{R}^{m \times d} \mid x_{\mathrm{fMRI}})$ of latent representations that capture underlying neural dynamics, where $m$ denotes the number of channels and $d$ is the feature dimension. 
A linear projection layer subsequently aggregates the channel-wise representations into a compact latent vector $z_c \in \mathbb{R}^{1 \times d}$, which will be passed to the decoder $D_n$ for neural signal reconstruction $\hat{x}_{\mathrm{fMRI}}=D_n(z_c)$.
Here, \textit{\textbf{$z_c$ serves to decouple the encoding pathway from the decoding process}}, enabling task-specific abstraction for fMRI synthesis.
Compared with SynBrain~\cite{mai2025synbrain}, an encoding-only method, we introduce a compact latent representation $z_c$, a cycle-consistency loss $\mathcal{L}_{\mathrm{cyc}}$, and several architectural modifications for bidirectional modeling  (see Appendix~\ref{sec:architecture} for details).

\vspace{-2mm}
\paragraph{Training Objectives.}
To encourage a probabilistic and semantically structured latent space alongside faithful fMRI reconstruction, NeuroVAE is optimized with a composite objective comprising the following objective terms.

\noindent\textbf{i) Reconstruction Loss} measures the voxel-wise \textit{mean squared error} between reconstructed fMRI signals $\hat{x}_{\mathrm{fMRI}}$ and original inputs $x_{\mathrm{fMRI}}$ to enforce voxel-level fidelity:
\begin{equation}
\mathcal{L}_{mse} = \lVert \hat{x}_{\mathrm{fMRI}} - x_{\mathrm{fMRI}} \rVert_2^2.
\end{equation}

\noindent\textbf{ii) KL Divergence Loss} regularizes the learned posterior distribution $q(z_n \mid x_{\mathrm{fMRI}})$ to be close to a standard Gaussian distribution $\mathcal{N}(0, I)$. This term introduces noises and encourages smoothness in the latent space to \textbf{\textit{model neural variability and facilitate continuous flow matching}} between neural and visual distributions, achieved by:
\begin{equation}
\mathcal{L}_{kl} = {\text{KL}}\left(q (z_n \mid x_{\text{fMRI}}) \parallel \mathcal{N}(0, I))\right..
\end{equation}

\noindent\textbf{iii) Contrastive Loss} facilitates cross-modal alignment by learning shared representations across modalities. This alignment encourages the neural latent space to encode semantic information that is consistent with visual stimuli, achieved by minimizing \textbf{SoftCLIP} loss~\citep{softclip}:
\begin{equation}
\mathcal{L}_{clip} = \text{SoftCLIP}(z_n,z_v),  \quad z_n = E_n(x_{\text{fMRI}}).
\end{equation}

\noindent\textbf{iv) Cycle-consistency Loss} ensures that reconstructed fMRI signals maintain semantic information instead of overfitting to the voxel-level fine-grained details, achieved by feeding synthetic signals into the neural encoder and computing \textbf{SoftCLIP} loss with visual representations:
\begin{equation}
\mathcal{L}_{cyc} = \text{SoftCLIP}(\hat{z}_n, z_v),  \quad \hat{z}_n = E_n(\hat{x}_{\text{fMRI}}).
\end{equation}

The final training loss is defined as a weighted sum of the above components:
\begin{equation}
\mathcal{L}_{\text{VAE}} = \mathcal{L}_{mse} + \alpha \mathcal{L}_{kl} + \beta \mathcal{L}_{clip} +
\lambda \mathcal{L}_{cyc}.
\end{equation}
% In our design, $\mathcal{L}_{kl}$ and $\mathcal{L}_{clip}$ are applied to the latent variable $z_n$ with \textbf{complementary purposes}: $\mathcal{L}_{kl}$ encourages smoothness and stochastic sampling capability in the latent space, while $\mathcal{L}_{clip}$ promotes semantic alignment between neural and visual distributions.

Here we set $\alpha = 0.001$ to \textbf{\textit{softly}} \textbf{\textit{regularize}} the latent distribution \textit{rather than strictly enforcing a standard Gaussian prior}, and set $\beta = 1,000$, $\lambda = 1,000$ to encourage a semantically organized latent space as well as semantically coherent fMRI reconstruction.
Through this design, conflicting objectives could be transformed into complementary ones. Specifically, NeuroVAE achieves a balance between $\mathcal{L}_{clip}$ and $\mathcal{L}_{kl}$ to prompt semantic structure while preserving stochastic sampling capability, and balances $\mathcal{L}_{mse}$ and $\mathcal{L}_{cyc}$ to prompt semantic fidelity while preserving \textit{critical} voxel-wise details.
Together, NeuroVAE provides a probabilistic and structured latent space for subsequent cross-model alignment and semantic-level fMRI synthesis.

% \vspace{-1mm}
\subsection{Cross-Modal Flow Matching (XFM)}
\vspace{-1mm}

\paragraph{Motivation. }
Visual encoding and decoding serve as complementary processes that model neural and visual distributions in the opposite direction. However, current models are biased toward the conditional noise-to-data diffusion strategy that synthesizes the target modality from noise distributions guided by another modality~\cite{bao2025mindsimulator,scotti2024mindeye2,scotti2023reconstructing}, deriving two issues: (i) \textit{Unidirectional modeling}: this strategy only builds a stochastic correspondence between one empirical distribution and Gaussian noise;
%, resulting in an inherently asymmetric mapping between neural and visual distributions
(ii) \textit{Training-inference distribution gap}: this strategy starts iterative denoising from a noisy empirical distribution during training, but inferences from pure Gaussian noise that is far from the training ones~\cite{mai2025synbrain}.
To tackle these issues, we propose Cross-model Flow Matching (XFM) to unify encoding and decoding processes by learning a reversibly consistent flow between empirical visual and neural latent distributions. 
% Diffusion models achieve approximate reversibility by pairing a forward noising process with a reverse denoising trajectory, establishing a stochastic correspondence between data and Gaussian noise. In cross-modal scenarios, however, one modality (e.g., fMRI) is typically serves as a condition that guides the denoising process toward another modality (e.g., images), resulting in an inherently asymmetric mapping. 

\vspace{-4mm}
\paragraph{Framework. }
For the first time, visual encoding and decoding are reformulated as \textbf{\textit{a time-dependent, reversible process within a shared latent space}}, as illustrated in Fig.~\ref{fig:overview}-B.
This formulation derives \textit{reversibility} from the uniqueness of ordinary differential equation (ODE) solutions: the learned vector field can be integrated forward for visual encoding $z_{\text{v}} \rightarrow z_{\text{n}}$ or backward for visual decoding $z_{\text{n}} \rightarrow z_{\text{v}}$. 
Given that, \textit{\textbf{encoding-decoding consistency}} is rigorously enforced by principles of flow matching.
XFM simulates the biological encoding process and regards the decoding process as a reversible transformation.
This formulation adheres to the Bayesian relationship between encoding and decoding~\cite{naselaris2011encoding}, where neural distributions can be regarded as the likelihood of visual representations, shaped by the statistical regularities of visual stimuli, from which posterior visual distributions can be inferred.

Formally, we define a time-dependent vector field $v_\theta(z, t)$ that transports samples between neural and visual distributions:
\begin{equation}
    \frac{dz(t)}{dt} = v_\theta(z_t, t), 
    \quad z_0 = z_{\text{v}}, \ z_1 = z_{\text{n}}.
\end{equation}
The vector field is parameterized by a Scalable Interpolant Transform (SiT)~\cite{ma2024sit} backbone with positional and temporal embeddings. 
Following the theory of flow matching~\citep{lipman2023flow}, the intermediate state can be defined via \textbf{\textit{cosine interpolation}} between distributions:
\begin{equation}
    z_t = \alpha_t z_0 + \sigma_t z_1, \   
    \alpha_t = \cos^2\!\left(\tfrac{\pi}{2} t \right), \ \sigma_t = \sin^2\!\left(\tfrac{\pi}{2} t \right),
\end{equation}
and the corresponding target vector field is:
\begin{equation}
    v^\ast(z_t, t) = \tfrac{d\alpha_t}{dt} \cdot z_0 + \tfrac{d\sigma_t}{dt} \cdot z_1.
\end{equation}
The training objective minimizes the squared error between predicted and target fields under \textit{\textbf{uniform time sampling}}:
\begin{equation}
    \mathcal{L}_{\text{XFM}} 
    = \mathbb{E}_{t \sim \mathcal{U}(0,1)} \left[ 
        \lVert v_\theta(z_t, t) - v^\ast(z_t, t) \rVert_2^2 
    \right].
\end{equation}

% \subsection{Visual Encoding and Decoding Inference}
During inference, cross-modal translation is achieved by numerically solving the learned ODE with Euler updates parameterized by the learned vector field $v_\theta(z_t, t)$.
The update rule is given by:
\begin{equation}
    z_{t+\Delta t} = z_t + \Delta t \, v_\theta(z_t, t),
\end{equation}
where the sign of $\Delta t$ determines the inference direction. A positive $\Delta t$ integrates forward in time ($t_0 \!\to\! t_1$), performing \textbf{visual encoding} ($z_{\text{v}} \!\to\! z_{\text{n}}$), while a negative $\Delta t$ integrates backward in time ($t_1 \!\to\! t_0$), performing \textbf{visual decoding} ($z_{\text{n}} \!\to\! z_{\text{v}}$).  
In this way, NeuroFlow achieves a unified formulation for visual encoding and decoding, with the two processes distinguished solely by the temporal sampling direction, as shown in Fig.~\ref{fig:overview}-C.

% \subsection{Visual Encoding and Decoding Inference}
% In inference, visual encoding and decoding can be done within a single model 

\vspace{-1mm}
\section{Experiments}
\vspace{-1mm}

\textbf{Dataset.   }We perform experiments on the Natural Scenes Dataset (NSD)~\cite{nsd}, a large-scale fMRI dataset where eight participants viewed natural images from COCO~\cite{COCO} over approximately 40 hours of scanning. We restrict our analysis to four subjects (Sub-1, Sub-2, Sub-5, Sub-7) who completed all experimental sessions, following the standard protocols in this field. For each subject, 9,000 unique images are used for training, and evaluation is performed on a shared set of 1,000 test images, each presented in three trials to simulate neural variability. Additional dataset details are provided in
the Appendix~\ref{sec:NSD}.

 % \noindent
\vspace{-4mm}
\paragraph{Training Details.}
 Our model is implemented on a single NVIDIA A100-40G GPU, with all training completed within 5 hours. Specifically, \textbf{NeuroVAE} is trained using the AdamW optimizer over 50 training epochs, with hyperparameters set as follows: $(\beta_1, \beta_2) = (0.9, 0.999)$, $learning\_rate = 1 \times 10^{-4}$, $weight\_decay=0.05$, and $batch\_size=64$. The \textbf{XFM} is optimized under the same hyperparameter configuration and trained for 50,000 steps.

\begin{figure*}[!t]
\begin{center}
\centerline{\includegraphics[width=0.9\textwidth]{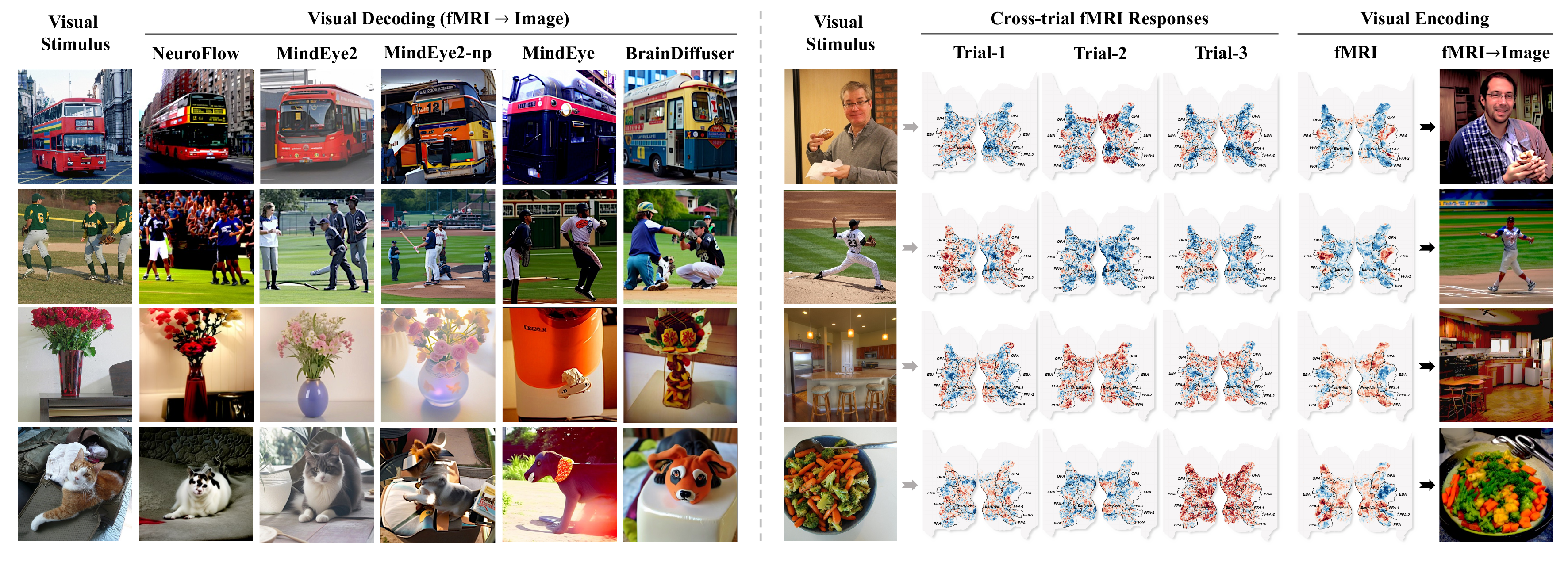}}
\vspace{-3mm}
\caption{\textbf{Qualitative visual encoding and decoding performance comparisons.} \textbf{Left:} NeuroFlow achieves superior decoding quality in semantic fidelity and visual structure. \textbf{Right:} NeuroFlow suppresses irrelevant cortical activity while enhancing category-specific regions, capturing consistent activation patterns underlying neural variability to support image synthesis consistent with visual stimuli. Positive (red) and negative (blue) values indicate increased and reduced activations, with color intensity reflecting the absolute magnitude of deviation.
% Positive (red) and negative (blue) values indicate increased and reduced activations, centered at zero, with color intensity reflecting the absolute magnitude of deviation.
}
\label{fig:main_results}
\end{center}
\vskip -0.3in
\end{figure*}

\begin{table*}[t]
    \centering
    % \captionsetup{font=small}
    \setlength{\tabcolsep}{4pt}
    \caption{Quantitative visual encoding and decoding performance comparisons.}
    \vspace{-2mm}
    \resizebox{0.9\textwidth}{!}{
    \begin{tabular}{lccccccccccc}
        \toprule
        \multirow{2}[1]{*}{Method} & \multirow{2}[1]{*}{Type} & \multicolumn{4}{c}{Decoding} & \multicolumn{4}{c}{Encoding ($\to$ Decoding)} & \multicolumn{2}{c}{Retrieval}\\
        \cmidrule(lr){3-6} \cmidrule(l){7-10} \cmidrule(l){11-12}
         & & Incep$\uparrow$ & CLIP$\uparrow$ & Eff$\downarrow$ & SwAV$\downarrow$ & Incep$\uparrow$ & CLIP$\uparrow$ & Eff$\downarrow$ & SwAV$\downarrow$ & Raw$\uparrow$ & Syn$\uparrow$ \\
         \midrule
        % MindSimulator~\cite{bao2025mindsimulator} & E & - & - & - & - & 92.1\% & 90.4\% & .701 & .396 & - & - \\
        MindSimulator~\cite{bao2025mindsimulator} & E & - & - & - & - & 93.1\% & 91.2\% & .689 & .391 & - & - \\
        SynBrain~\cite{mai2025synbrain} & E & - & - & - & - & 95.7\% & 94.3\% & .639 & .362 & 84.8\% & 92.5\%\\
        % \midrule
        BrainDiffuser~\cite{ozcelik2023natural} & D & 91.3\% & 90.9\% & .728 & .421  & - & - & - & - &  18.8\% & -\\
        MindEye~\cite{scotti2023reconstructing} & D & 94.6\% & 93.3\% & .622 & \textbf{.343}  & - & - & - & - & 90.0\% & -\\
        MindEye2~\cite{scotti2024mindeye2} & D & 95.4\% & 93.0\% & \textbf{.618} & .344 & - & - & - & - & \textbf{98.8\%} & -\\
        \midrule
        NeuroFlow & E\&D & \textbf{95.6\%} & \textbf{94.2\%} & .674 & .364 & \textbf{98.6\%} & \textbf{98.7\%} & \textbf{.590} & \textbf{.341} & 80.6\% & \textbf{97.0\%} \\
        \bottomrule
    \end{tabular}
    }
    \label{tab:main_results}
    \vspace{-4mm}
\end{table*}

 % \noindent
\vspace{-4mm}
\paragraph{Evaluation Metrics.}
We measure visual encoding and decoding performance at the \textit{semantic level}:

\noindent\textbf{i) Visual Decoding (fMRI$\rightarrow$Image).}  
Given fMRI signals, we generate images and assess their semantic fidelity against the original visual stimuli using Inception Score (Incep)~\cite{inception}, CLIP similarity (CLIP)~\cite{clip}, EfficientNet distance (Eff)~\cite{eff}, and SwAV distance (SwAV)~\cite{swav}.

\noindent\textbf{ii) Visual Encoding (Image$\rightarrow$fMRI$\rightarrow$Image).} Same as the decoding metrics above, but focuses on \textbf{\textit{encoding-decoding consistency}} that assesses whether the neural representation of synthetic signals preserves critical semantic information for faithful image synthesis.

\noindent\textbf{iii) Retrieval Metrics.} 
We evaluate how well fMRI signals preserve semantic information by computing top-1 retrieval accuracy based on cosine similarity between neural ($z_n$$/\hat{z}_n$) and visual ($z_v$) latent representations \cite{scotti2024mindeye2,mai2025synbrain}. Two signal sources are compared: (i) raw fMRI (Raw) $\to z_n$, and (ii) synthetic fMRI signals (Syn) $\to \hat{z}_n$. 
% Retrieval results are averaged over 30 sampled subsets of 300 candidate images, following MindEye2~\cite{scotti2024mindeye2} and SynBrain~\cite{mai2025synbrain}.

Together, these metrics provide a comprehensive assessment of semantic consistency across modalities, covering visual encoding, decoding, and encoding–decoding consistency. Details are provided in the Appendix~\ref{sec:Metric}.

\begin{table}[t]
    \centering
    \setlength{\tabcolsep}{4pt}
    \caption{Model efficiency comparisons.}
    \vspace{-2mm}
    \resizebox{0.47\textwidth}{!}{
    \begin{tabular}{lcccc}
        \toprule
        Method & Type & Pretrain & Architecture & Param \\
        \midrule
        SynBrain~\cite{mai2025synbrain} & E & $\boldsymbol{\times}$ & VAE+Transformer & 690M  \\
        % BrainDiffuser~\cite{ozcelik2023natural} & D & $\boldsymbol{\times}$ & Linear & 3.08G \\
        MindEye~\cite{scotti2023reconstructing} & D & $\boldsymbol{\times}$ & MLP+DP & 1.00B \\
        MindEye2~\cite{scotti2024mindeye2} & D & $\boldsymbol{\checkmark}$ & Linear+MLP+DP & 2.60B  \\
        \midrule
        NeuroFlow & E\&D & $\boldsymbol{\times}$ & VAE+XFM & 660M  \\
        \bottomrule
    \end{tabular}
    }
    \label{tab:efficiency}
    \vskip -0.2in
\end{table}

\begin{figure*}[!t]
\begin{center}
\centerline{\includegraphics[width=0.9\textwidth]{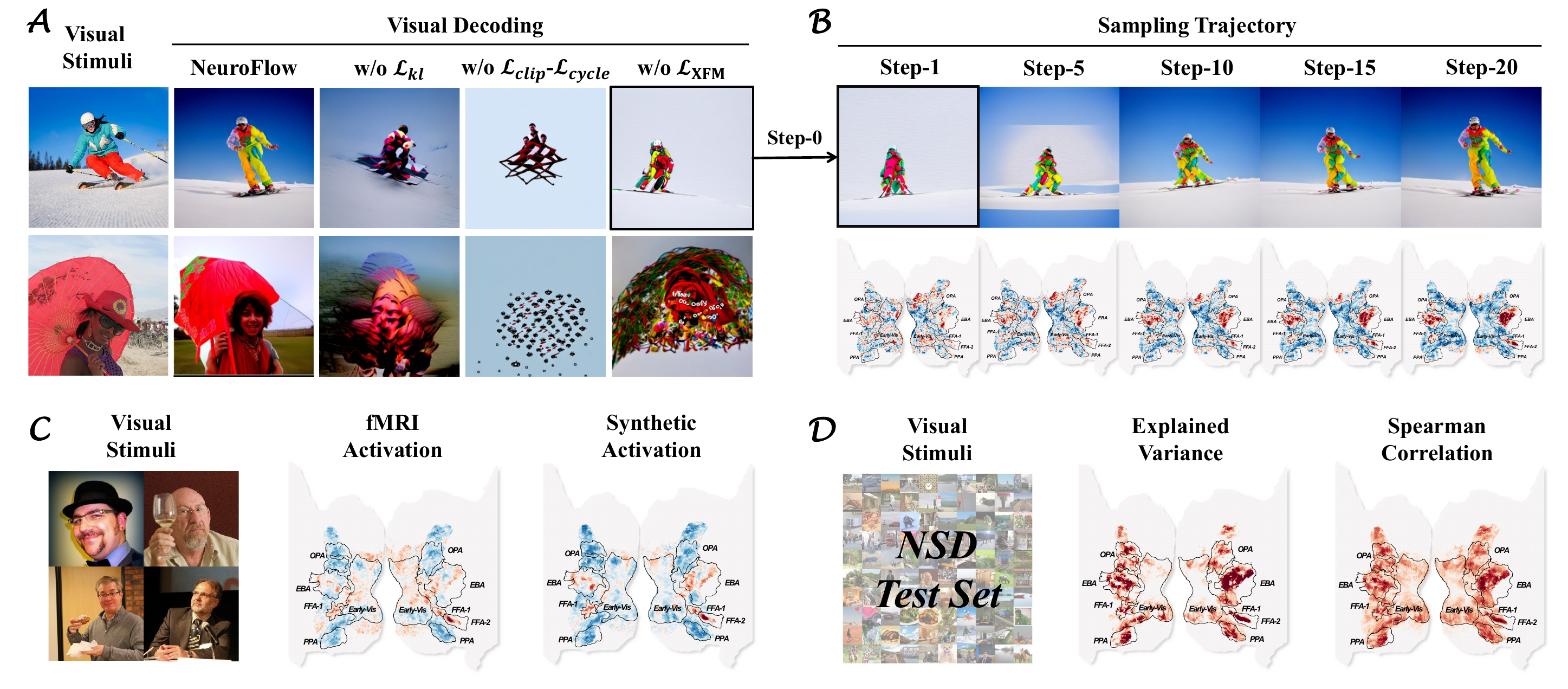}}
\vspace{-3mm}
\caption{\textbf{Empirical visualizations. (A)} \textbf{Ablation study}: removing key objectives leads to degraded visual fidelity and semantic coherence.
\textbf{(B)} \textbf{Flow trajectory}: Encoding trajectory reveals a suppression of early visual responses and a transition toward functional regions (i.e., FFA and EBA). Decoding trajectory evolves from an initial structural sketch, \textit{not Gaussian noises}, to a realistic and high-fidelity image. \textbf{(C-D) Brain functional analysis}: category-selective fMRI activations and voxel-wise evaluation derived from raw and synthetic fMRI, computed on the whole test set, showing that NeuroFlow suppresses early visual activity and emphasizes higher-order functional regions.
}
\label{fig:sampler}
\end{center}
\vskip -0.3in
\end{figure*}

\begin{table*}[t]
    \centering
    % \captionsetup{font=small}
    \setlength{\tabcolsep}{5pt}
    \caption{Quantitative performance comparisons between NeuroFlow and its ablated variants on Subject 1.}
    \vspace{-2mm}
    \resizebox{0.85\textwidth}{!}{
    \begin{tabular}{lcccccccccc}
        \toprule
        \multirow{2}[1]{*}{Method} & \multicolumn{4}{c}{Decoding} & \multicolumn{4}{c}{Encoding ($\to$ Decoding)} & \multicolumn{2}{c}{Retrieval}\\
        \cmidrule(lr){2-5} \cmidrule(l){6-9} \cmidrule(l){10-11}
         & Incep$\uparrow$ & CLIP$\uparrow$ & Eff$\downarrow$ & SwAV$\downarrow$ & Incep$\uparrow$ & CLIP$\uparrow$ &Eff$\downarrow$ & SwAV$\downarrow$ & Raw$\uparrow$ & Syn$\uparrow$ \\
        \midrule
        w/o $z_c$ & 94.7\% & 94.0\% & .683 & .378 & 97.2\% & 96.6\% & .636 & .373 & 84.1\% & 90.4\%\\
        w/o $\mathcal{L}_{kl}$ & 89.0\% & 79.6\% & .807 & .475 & 60.7\% & 57.1\% & .956 & .606 & 75.1\% & 11.5\%\\
        w/o $\mathcal{L}_{cyc}$ & 94.9\% & 93.7\% & .687 & .383 & 96.8\% & 95.5\% & .658 & .382 & 81.4\% & 88.9\%\\
        w/o $\mathcal{L}_{clip}$-$\mathcal{L}_{cyc}$ & 57.6\% & 58.8\% & .961 & .674 & 52.6\% & 51.3\% & .983 & .700 & 0.3\% & 0.5\%\\
        w/o $\mathcal{L}_\text{XFM}$ & 87.7\% & 83.7\% & .788 & .497 & 60.3\% & 58.1\% & .953 & .644 & 86.4\% & 14.1\%\\
        \midrule
        NeuroFlow & \textbf{95.9\%} & \textbf{95.0\%} & \textbf{.675} & \textbf{.370} & \textbf{98.5\%} & \textbf{98.7\%} & \textbf{.600} & \textbf{.347} & \textbf{86.4\%} & \textbf{96.4\%}\\
        \bottomrule
    \end{tabular}
    }
    \label{tab:ablation}
    \vspace{-5mm}
\end{table*}

\vspace{-1mm}
\section{Results and Analysis}
\vspace{-1mm}
\subsection{Main Results}
\vspace{-1mm}

We evaluate \textbf{NeuroFlow} on visual encoding and decoding tasks, providing qualitative visualizations and quantitative results compared with state-of-the-art methods.

\vspace{-4mm}
\paragraph{Visual Decoding.}
NeuroFlow achieves comparable performance with MindEye2, while significantly surpassing other methods in the coherence of semantic fidelity and visual structure, as shown in Fig.~\ref{fig:main_results}-Left. 
Here, MindEye2 (np) is a variant of MindEye2 that is \textit{not pretrained}, offering a fairer comparison with our approach. 
Even though \textbf{\textit{no explicit low-level visual information}} is incorporated into NeuroFlow, distinct from all compared methods, our approach can still generate coherent structures consistent with the original stimuli. We attribute this capability to the visual decoder in NeuroFlow, which is an UnCLIP model that trains to faithfully reproduce the reference images~\cite{scotti2024mindeye2}.
Quantitatively, NeuroFlow outperforms all decoding baselines, reaching the best overall performance in \textbf{Incep} (95.6\%) and \textbf{CLIP} (94.2\%) scores, while maintaining competitive results in Eff and SwAV distances.
These results demonstrate that NeuroFlow effectively reproduces semantically coherent visual content from fMRI activity.

\vspace{-4mm}
\paragraph{Visual Encoding ($\to$ Decoding).}
Neural responses to identical stimuli vary across trials with substantial voxel-level variability, as illustrated in Fig.~\ref{fig:main_results}-Right. 
NeuroFlow effectively learns to \textit{abstract away fine-grained fluctuations and captures consistent patterns across trials}. 
Specifically, NeuroFlow selectively compresses early visual and irrelevant cortical regions, thereby enhancing activation in category-specific areas, such as the FFA (faces), EBA (bodies), OPA/PPA (places), and the food-selective region.
Despite fine-grained differences, our synthetic fMRI signals possess coherent activation patterns to reproduce images that are semantically consistent with the original stimuli.

Quantitatively, we assess whether synthetic fMRI signals retain essential semantic information to support coherent decoding. 
As shown in Tab.~\ref{tab:main_results}, NeuroFlow significantly surpasses all encoding models, exhibiting strong encoding-decoding consistency within a single model.
Although NeuroFlow sacrifices raw retrieval performance for unified modeling, its encoding and decoding performance remain superior.
Notably, decoding and retrieval performance using synthetic fMRI signals even outperform those based on raw fMRI (e.g., Incep: 98.6\% vs. 95.6\%, and Retrieval: 97.0\% vs. 80.6\%). 
These results indicate that NeuroFlow effectively distills task-relevant, semantic information from sparse and redundant fMRI signals and synthesizes neural signals that are more coherent with visual semantics.

\vspace{-4mm}
\paragraph{Model Efficiency Comparison.}
Tab.~\ref{tab:efficiency} summarizes model efficiency in terms of architecture, pretraining, and parameter scale. 
NeuroFlow serves as a unified framework for visual encoding and decoding, yet it comprises only \textbf{660M} trainable parameters without any pretraining--approximately \textbf{25\%} of the parameter count of \textit{MindEye2} (2.60B). 
Despite its compact design, it achieves performance comparable to, and in several metrics surpassing, that of MindEye and MindEye2.
% Moreover, it significantly exceeds all encoding models in both performance and efficiency, requiring only a fraction of their parameters. 
These results highlight the strong scalability and parameter efficiency of NeuroFlow, exhibiting that a unified framework can achieve superior performance with substantially reduced computational cost.

\vspace{-1mm}
\subsection{Ablation Study}
\label{sec:ablation}
\vspace{-1mm}
To investigate the contribution of key components in NeuroFlow, we conduct systematic ablation experiments on Subject 1, examining the effects of removing the compact latent ($z_c$), variational sampling ($\mathcal{L}_{kl}$), cycle-consistency mechanism ($\mathcal{L}_{cyc}$), contrastive learning ($\mathcal{L}_{clip}$-$\mathcal{L}_{cyc}$), and XFM ($\mathcal{L}_{\text{XFM}}$). Results are summarized in Tab.~\ref{tab:ablation}.

\noindent\textbf{Impact of Compact Latent ($z_c$). }Removing this term leads to consistent performance drops across tasks. As an encoding-specific branch derived from $z_n$, $z_c$ further isolates and compresses task-relevant information for fMRI synthesis. This separation of pathways, where $z_n$ supports decoding and $z_c$ supports encoding, enhances the overall consistency and effectiveness of bidirectional modeling.

\noindent\textbf{Impact of Variation Sampling ($\mathcal{L}_{kl}$).}  
Removing this term leads to severe performance degradation across all metrics, particularly in the encoding direction. This highlights the importance of the \noindent\textit{probabilistic latent space} for one-to-many cross-modal alignment (Raw Retrieval: 86.4\%$\to$75.1\%) and continuous cross-modal flow matching (Syn Retrieval: 96.4\%$\to$11.5\%) between visual and neural distributions. As shown in Fig.~\ref{fig:sampler}-A, the absence of $\mathcal{L}_{\text{KL}}$ leads to highly distorted synthetic images that lose semantic fidelity and structural coherence.

\noindent\textbf{Impact of Cycle-Consistency Mechanism ($\mathcal{L}_{cyc}$).}  
Removing this component results in moderate degradation in all metrics, indicating that enforcing synthetic fMRI signals to focus on semantic-aware patterns instead of overfitting to the voxel-level details enhances encoding/decoding stability and cross-modal coherence.

\noindent\textbf{Impact of Contrastive Learning ($\mathcal{L}_{clip}$–$\mathcal{L}_{cyc}$).}
NeuroFlow incorporates two contrastive objectives to align neural and visual latent representations. 
Removing them leads to a dramatic collapse in retrieval performance (Raw: $86.4\%\!\rightarrow\!0.3\%$, Syn: $96.4\%\!\rightarrow\!0.5\%$), highlighting the necessity of contrastive alignment for constructing a shared latent space across modalities. 
Beyond retrieval, contrastive learning also provides a \textbf{\textit{coarse but critical alignment} }between the neural and visual distributions, which serves as \textit{a prerequisite for effective cross-modal flow matching}. 
Without that, XFM fails to establish direct flows between the two distributions without any guidance, resulting in a drastic decline in encoding and decoding performance.
As shown in Fig.~\ref{fig:sampler}-A,  the absence of $\mathcal{L}_{\text{clip}}$–$\mathcal{L}_{\text{cyc}}$ leads to severe visual degradation, i.e, synthetic images lose semantic content and collapse into repetitive or texture-like artifacts. 

\noindent\textbf{Impact of Cross-Modal Flow Matching ($\mathcal{L}_{\text{XFM}}$).}  
XFM is critical for bridging the residual modality gap between neural and visual distributions that remains after contrastive alignment. 
Removing the XFM module yields substantial degradation across all metrics, most notably on the encoding side, indicating that XFM is the key mechanism that unifies visual encoding and decoding with strong semantic consistency.
As shown in Fig.~\ref{fig:sampler}-A, ablating $\mathcal{L}_{\text{XFM}}$ prevents the synthesis of realistic images, producing only distorted outlines and fragmented textures.
We further demonstrate that replacing XFM with a simple MSE objective (NeuroVAE+MSE) or two linear regressions (NeuroVAE+LRs) as baseline models leads to significant performance degradation, underscoring the importance of XFM for unified modeling (see Appendix~\ref{sec:baseline} for details).

\vspace{-1mm}
\subsection{Sampling Trajectory}
\vspace{-1mm}
Fig.~\ref{fig:sampler}-B illustrates the stepwise flow trajectory of NeuroFlow over 20 sampling steps, computed using the Euler solver.
Unlike diffusion models that initialize from Gaussian noise and rely on strong guidance to approach the target distribution, XFM directly learns continuous, reversible flows between the neural and visual latent distributions.
The encoding trajectory reveals a suppression of early visual responses and a gradual transition toward category-selective regions, e.g., EBA and FFA, corresponding to the stimulus with body and face.
The decoding trajectory is initialized from a semantically grounded neural distribution (NeuroFlow w/o $\mathcal{L}_{\text{XFM}}$, Step-0), with early frames capturing the visual structure and later iterations refining fine details to produce realistic and semantically coherent images. 
This strategy shortens the sampling path, stabilizes trajectories, and preserves coherent semantics throughout the sampling process, establishing a stable and reversible pathway between the two distributions without external guidance. 
% Simply reverse the direction to start from the visual distribution and approach the neural distribution for the encoding pathway.

\vspace{-1mm}
\subsection{Brain Functional Analysis}
\vspace{-1mm}
% To further understand the functional characteristics of the neural signals generated by NeuroFlow, we conduct a comprehensive brain-level analysis using category-selective fMRI activations and voxel-wise evaluation metrics.
We first analyze category-selective fMRI activations by comparing raw and synthetic neural responses across various stimulus categories. As shown in Fig.~\ref{fig:sampler}-C, NeuroFlow preserves the functional selectivity of cortical responses, with synthetic fMRI signals selectively activating regions (i.e., FFA and EBA). These results demonstrate that the model captures biologically meaningful patterns and maintains consistency with known cortical organization.
To further quantify the alignment between synthetic and raw neural signals, we compute voxel-wise \textit{Explained Variance (EV)} and \textit{Spearman Correlation} across the whole NSD test set (Fig.~\ref{fig:sampler}-D). The results reveal that NeuroFlow suppresses activity in early visual areas (e.g., V1–V4) and enhances signals in higher-order visual cortices, including FFA, EBA, and PPA. These patterns reveal that NeuroFlow prioritizes semantically relevant neural components and generates functionally aligned representations.
Overall, these findings confirm that NeuroFlow not only improves encoding-decoding performance but also produces neural signals that are interpretable and functionally aligned with human visual cortex organization.

\vspace{-1mm}
\section{Conclusions}
\vspace{-1mm}
We introduce \textbf{NeuroFlow}, the first unified framework that jointly models visual encoding and decoding from neural activity. By integrating a variational neural backbone (NeuroVAE) with cross-modal flow matching (XFM), NeuroFlow establishes a shared latent space and enforces encoding-decoding consistency in a principled, reversible manner. Empirical results demonstrate that NeuroFlow achieves superior overall performance across tasks with higher parameter efficiency. Further brain functional analyses show that NeuroFlow preserves biologically meaningful activations and suppresses irrelevant neural noise, producing interpretable and functionally aligned representations.
NeuroFlow takes a major step toward bridging encoding and decoding in a unified model and provides insights into future development of bidirectional, biologically grounded brain–computer interfaces.

\section*{Acknowledgment}
This work is supported by Shanghai Artificial Intelligence Laboratory. 
This work is supported by Intern Discovery.
This work was done during Weijian Mai's internship at Shanghai Artificial Intelligence Laboratory.

{
    \small
    \bibliographystyle{ieeenat_fullname}
    \bibliography{ref}
}

% WARNING: do not forget to delete the supplementary pages from your submission 
% \input{sec/X_suppl}

\clearpage
% \newpage 
\appendix

\section{NSD Dataset}
\label{sec:NSD}
In this study, we leverage the largest publicly available fMRI-image dataset, the Natural Scenes Dataset (NSD) \cite{nsd}, which encompasses extensive 7T fMRI data collected from eight subjects while they viewed images from the COCO dataset. Each subject viewed each image for 3 seconds and indicated whether they had previously seen the image during the experiment.
Our analysis focuses on data from four subjects (Sub-1, Sub-2, Sub-5, and Sub-7) who completed all viewing trials. The training dataset consists of 9,000 images and 27,000 fMRI trials, while the test dataset includes 1,000 images and 3,000 fMRI trials, with up to 3 repetitions per image. It is important to note that the test images are consistent across all subjects, whereas distinct training images are utilized.

We used preprocessed scans from NSD for functional data, with a resolution of 1.8 mm. Our analysis involved employing single-trial beta weights derived from generalized linear models, along with region-of-interest (ROI) data specific to early and higher (ventral) visual regions as provided by NSD. The ROI voxel counts for the respective four subjects are as follows: [15724, 14278, 13039, 12682].
% Detailed fMRI preprocessing procedures and additional information can be found in the source paper \cite{nsd} and on the NSD website\footnote{https://naturalscenesdataset.org}.

\subsection{Preprocessing}
We perform \textbf{per-session z-score normalization} to centre each voxel to zero mean and scale it to unit variance within its respective session.
All trials are retained individually for model training, while repeated presentations in the test set are averaged to yield a single, denoised beta pattern for each stimulus.
To focus on visual cortical processing, we employ the official \textbf{nsdgeneral} ROI mask, which spans early to higher-order visual regions. Voxels within this mask are extracted and flattened into a one-dimensional sequence that serves as the fMRI input $x_{\mathrm{fMRI}} \in \mathbb{R}^{1 \times n}$ to the neural encoder.

\subsection{Brain Functional Regions}
High-level visual cortex contains multiple \textbf{category-selective regions} that respond preferentially to distinct types of visual stimuli. In humans, the most extensively studied functional regions are those selective for \textbf{faces, bodies, places, and food}, which exhibit reliable and dissociable responses across individuals. These regions provide a well-characterized framework for investigating category-specific neural representations and are commonly targeted in fMRI studies of visual processing.

\textbf{Face-selective regions.} 
These areas respond preferentially to faces and primarily include the fusiform face area (FFA). They are organized along the ventral visual pathway and support the perception of facial identity and expression.

\textbf{Body-selective regions.} Located adjacent to face-selective cortex, these regions respond strongly to images of human bodies, with the extrastriate body area (EBA) playing a key role in encoding body form, posture, and movement.

\textbf{Place-selective regions.} This network responds most strongly to scenes and environmental layouts. It comprises the parahippocampal place area (PPA) and the occipital place area (OPA), which collectively encode spatial structure and navigationally relevant information.

\textbf{Food-selective regions.} These regions are located within the ventral temporal cortex and show enhanced responses to edible items and food-related visual features.

\begin{figure}[!t]
\begin{center}
\centerline{\includegraphics[width=0.5\textwidth]{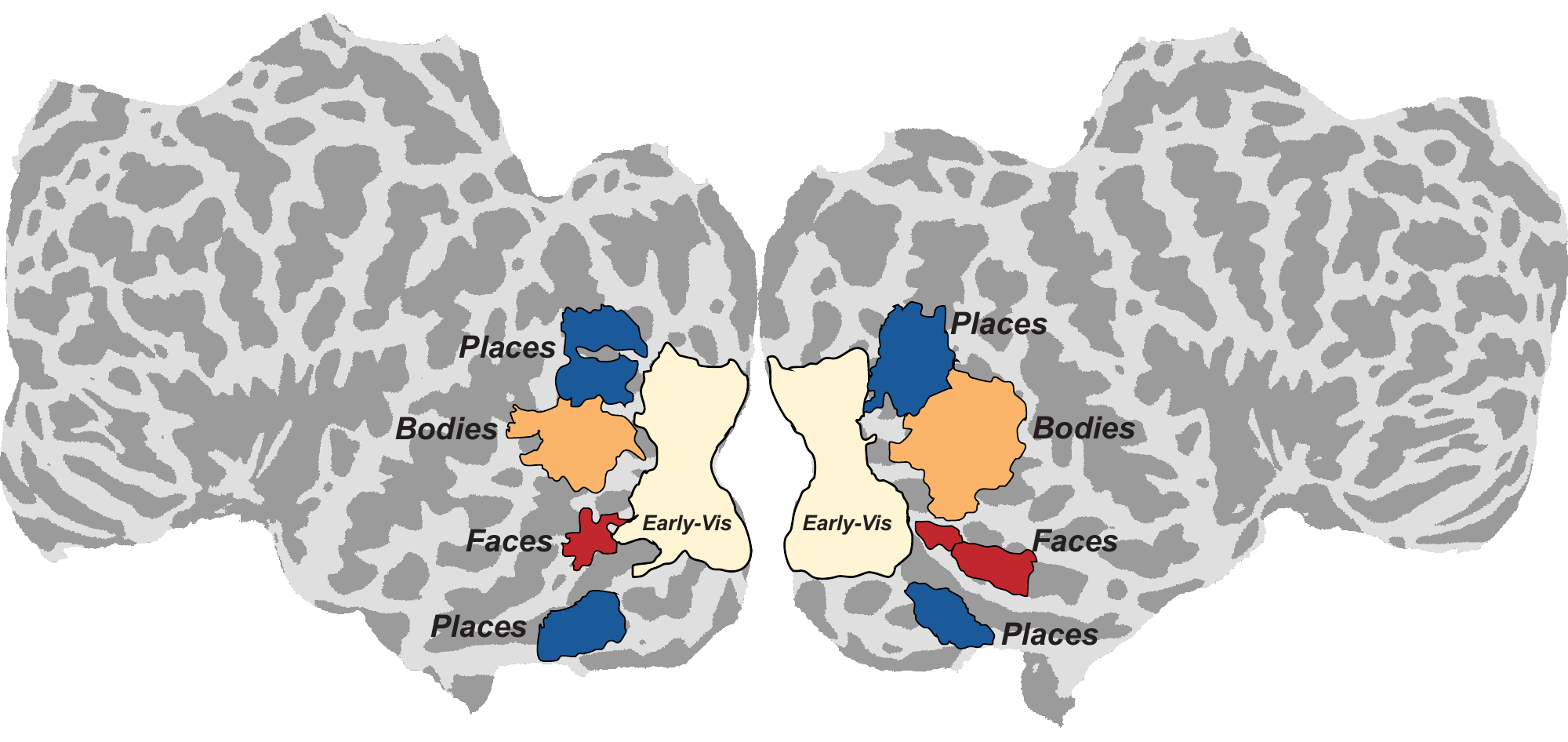}}
\vspace{-3mm}
\caption{High-order functional regions in the visual cortex. 
}
\label{fig:brain_region}
\end{center}
\vskip -0.3in
\end{figure}

\section{Evaluation Metric}
\label{sec:Metric}
We evaluate visual encoding and decoding at the \textbf{semantic level} rather than at the voxel or pixel level, as lower-level analyses exhibit poor consistency between encoding and decoding, leading to blurry reconstructions.

\vspace{-4mm}
\paragraph{i) Visual Decoding (fMRI$\rightarrow$Image).  }
Given fMRI signals, we generate images and assess their semantic fidelity against the original visual stimuli using multiple semantic metrics: i) \textbf{Incep:} A two-way comparison of the last pooling layer of InceptionV3; ii) \textbf{CLIP:} A two-way comparison of the output layer of the CLIP-Image model; iii) \textbf{Eff:} A distance metric gathered from EfficientNet-B1 model; iv) \textbf{SwAV: }A distance metric gathered from SwAV-ResNet50 model.
A two-way comparison evaluates the accuracy percentage by determining whether the original image embedding aligns more closely with its corresponding brain embedding or with a randomly selected brain embedding.

\vspace{-4mm}
\paragraph{ii) Visual Encoding (Image$\rightarrow$fMRI$\rightarrow$Image). }
Same as the decoding metrics above, but focuses on \textbf{\textit{encoding-decoding consistency}} that assesses whether the neural representation of synthetic signals preserves critical semantic information for faithful image synthesis. 
Note that the encoding-only models (i.e., SynBrain~\cite{mai2025synbrain} and MindSimulator~\cite{bao2025mindsimulator}) need to employ the decoding-only model (i.e., MindEye2~\cite{scotti2024mindeye2}) as the frozen fMRI-to-image generator to evaluate the semantic quality of synthetic fMRI signals.

\vspace{-4mm}
\paragraph{iii) Retrieval Metrics. }
We evaluate how well fMRI signals preserve semantic information by computing top-1 retrieval accuracy based on cosine similarity between neural latent representations ($z_n$$/\hat{z}_n$) and 300 candidate visual latent representations ($z_v$) extracted from test images, with one being the ground-truth visual stimulus for the fMRI data \cite{scotti2024mindeye2,mai2025synbrain}. Two signal sources are compared: (i) raw fMRI (Raw) $\to z_n$, and (ii) synthetic fMRI signals (Syn) $\to \hat{z}_n$. 
Retrieval performance is evaluated by calculating the average Top-1 retrieval accuracy (with a chance level of 1/300) and repeating the process 30 times to account for batch sampling variability.

Together, these metrics offer a comprehensive evaluation of semantic consistency across modalities, encompassing visual encoding, decoding, and the consistency between encoding and decoding.

\begin{algorithm}[t]
\caption{NeuroVAE Architecture}
\label{alg:neurovae}
\begin{algorithmic}[1]
\State \textbf{Input:} fMRI signal $x \in \mathbb{R}^{1 \times n} $
% , visual representation $z_{\text{v}}$
\State \textbf{Encoder:}
    \State \hspace{1em} $x \gets \texttt{Conv1D}(x,\ c\_out=64)$
    \State \hspace{1em} $x \gets \texttt{MLP}(x,\ h=1024,\ d\_out=6656)$
    \State \hspace{1em} $x \gets \texttt{DownBlock}(x,\ num\_block=2)$

\State \textbf{Sampling:}
\State \hspace{1em} $[\mu,\ \log\sigma^2] = \texttt{Conv1d}(x,\ c\_out=512)$
\State \hspace{1em} $z_n \sim \mathcal{N}(\mu,\ \sigma^2)$  \Comment{$z_n \in \mathbb{R}^{256 \times 1664} $}

\State \textbf{Pre-Projector:}
\State \hspace{1em} $z_c = \texttt{Conv1d}(z_n,\ c\_out=1)$  \Comment{$z_c \in \mathbb{R}^{1 \times 1664} $}
\State \textbf{Post-Projector:}
\State \hspace{1em} $x = \texttt{Conv1d}(z_c,\ c\_out=256)$

\State \textbf{Decoder:}
    \State \hspace{1em} $x \gets \texttt{UpBlock}(x,\ num\_block=2)$
    \State \hspace{1em} $x \gets \texttt{MLP}(x,\ h=1024,\ d\_out=n)$
    \State \hspace{1em} $\hat{x} \gets \texttt{Conv1D}(x,\ c\_out=1)$
\State \Return $\hat{x} \in \mathbb{R}^{1 \times n}$
\end{algorithmic}
\end{algorithm}

\section{Architecture Details}
\label{sec:architecture}
\subsection{NeuroVAE Architecture}

The overall architecture of NeuroVAE is summarized in Algorithm~\ref{alg:neurovae}. Given an fMRI input $x_{\mathrm{fMRI}} \in \mathbb{R}^{1 \times n}$, the encoder first applies a $1 \times 1$ Conv1D, followed by an MLP and two downsampling blocks to produce intermediate features. A Conv1D layer then estimates the posterior parameters $[\mu, \log\sigma^2]$ for the latent representation $z_n \sim \mathcal{N}(\mu, \sigma^2)$. A pre-projector Conv1D compresses $z_n$ into a channel-aggregated vector $z_c \in \mathbb{R}^{1 \times d}$, which is mapped back through a post-projector Conv1D and two upsampling blocks, followed by an MLP and final Conv1D, to reconstruct the fMRI signal $\hat{x}_{\mathrm{fMRI}}$. Here, $c\_out$ denotes the output number of channels, and $d\_out$ denotes the output number of feature dimensions.
This design decouples the encoding and decoding pathways, resulting in a compact, probabilistic latent space that captures essential neural dynamics while supporting bidirectional cross-modal generative modeling.

Building on this foundation, NeuroVAE provides a variational backbone that constructs a semantically organized latent space for mapping between fMRI and visual embeddings. Compared with SynBrain~\cite{mai2025synbrain}, NeuroVAE introduces several key improvements motivated by biological plausibility, computational efficiency, and the requirements of generative neural modeling.

\textbf{i) Biologically motivated feature processing.} 
The encoder processes one-dimensional fMRI inputs using $1 \times 1$ Conv1D layers, which act as \textit{channel-wise linear transformations} rather than spatial convolutions. Since \textbf{\textit{voxel ordering in the 1D flattened fMRI vector does not encode meaningful spatial relationships}}, this design avoids injecting artificial spatial inductive biases. SynBrain instead applies adaptive max pooling, which implicitly assumes spatial locality; NeuroVAE replaces this with an MLP projection to obtain a more biologically justified global transformation.

\textbf{ii) Reduced attention dimensionality and improved computational efficiency.} 
SynBrain computes attention over a $\mathbb{R}^{512 \times 4096}$ representation, requiring four A100-40GB GPUs for training. NeuroVAE reduces the representation to $\mathbb{R}^{256 \times 1664}$, which aligns with the CLIP visual feature space and allows training on a single A100-40GB GPU. This reduction preserves semantic alignment while substantially lowering memory and compute demands.

\textbf{iii) Compact latent space enabling bidirectional modeling.} 
SynBrain retains a $\mathbb{R}^{256 \times 1664}$ latent tensor and supports only encoding. NeuroVAE aggregates channel-wise information into a single latent vector $z_c \in \mathbb{R}^{1 \times 1664}$, cleanly separating the encoder and decoder pathways. This compact latent representation enables both neural reconstruction and generative fMRI synthesis through cross-modal flow matching.

\textbf{iv) Cycle-consistency loss for semantically coherent fMRI synthesis.} 
To ensure that reconstructed fMRI signals maintain semantic fidelity, NeuroVAE introduces \textbf{\textit{a cycle-consistency loss}} that feeds synthetic signals back into the encoder and aligns their latent representations with the corresponding visual embeddings. This encourages generative fMRI signals to preserve semantic information instead of overfitting to voxel-level noise.

Together, these improvements allow NeuroVAE to construct a structured and probabilistic neural latent space that is both computationally efficient and better suited for semantic-level bidirectional modeling between visual and neural domains.

\subsection{XFM Architecture}

The Cross-modal Flow Matching (XFM) module is built on a SiT backbone~\cite{ma2024sit} with temporal and positional embeddings. We use an 12-layer Transformer with 13 attention heads per layer. The neural and visual latent representations share the same dimensionality, with 256 tokens and a feature dimension of 1664 (i.e., $z_n, z_v \in \mathbb{R}^{256 \times 1664}$), which is required for \textit{\textbf{cosine interpolation}} when defining the continuous path between the two distributions. The patching and unpatching layers are removed since XFM operates directly in the high-dimensional latent space.
% and 2D positional embeddings are replaced with a 1D variant due to the absence of spatial structure in the tokens.

Bypassing the Gaussian noise distributions in standard implementations, we treat visual and neural distributions as the initial ($z_0,t=0$) and target distributions($z_1,t=1$), and learn a reversibly consistent flow between them. 
This reframes the unidirectional denoising process into a unified cross-modal transport. We further remove the classifier guidance to facilitate a direct transport between the two distributions that aligns more closely with the biological process of visual encoding and decoding.
For the first time, we reformulate visual encoding and decoding as \textbf{\textit{a time-dependent, reversible process}} for unified modeling between neural and visual latent distributions. 
This formulation derives \textit{\textbf{reversibility}} from the uniqueness of ordinary differential equation (ODE) solutions: the learned vector field can be integrated forward for visual encoding $z_{\text{v}} \rightarrow z_{\text{n}}$ or backward for visual decoding $z_{\text{n}} \rightarrow z_{\text{v}}$. 
Given that, \textit{\textbf{encoding-decoding consistency}} is rigorously enforced by principles of flow matching.
During inference, cross-modal translation is achieved by numerically solving the learned ODE with Euler updates parameterized by the learned vector field. In this way, NeuroFlow achieves a unified formulation for visual encoding and decoding, with\textbf{\textit{ the two processes distinguished solely by the temporal sampling direction}}.

\paragraph{Ablation Study on Interpolation Schedule.}
We assessed the effect of different interpolation schedules on cross-modal flow performance. As reported in Tab.~\ref{tab:interpolation}, NeuroFlow performs well under both linear and cosine schedules, demonstrating the robustness of our approach. The cosine schedule yields slightly better results, likely due to its smoother transition, which enables a more stable and accurate mapping between the visual and neural latent distributions. The smoother progression helps reduce abrupt changes in the latent space, facilitating better alignment and more reliable flow estimation. These findings indicate that while NeuroFlow is inherently robust, careful design of the interpolation path can further enhance performance.

\begin{table*}[t]
    \centering
    \setlength{\tabcolsep}{5pt}
    \caption{Ablation experiments on linear and cosine interpolation schedules.}
    \vspace{-2mm}
    \resizebox{0.9\textwidth}{!}{
    \begin{tabular}{lcccccccccc}
        \toprule
        \multirow{2}[1]{*}{Schedule} & \multicolumn{4}{c}{Decoding} & \multicolumn{4}{c}{Encoding ($\rightarrow$Decoding)} & \multicolumn{2}{c}{Retrieval}\\
        \cmidrule(lr){2-5} \cmidrule(l){6-9} \cmidrule(l){10-11}
         & Incep$\uparrow$ & CLIP$\uparrow$ & Eff$\downarrow$ & SwAV$\downarrow$ & Incep$\uparrow$ & CLIP$\uparrow$ &Eff$\downarrow$ & SwAV$\downarrow$ & Raw$\uparrow$ & Syn$\uparrow$ \\
        \midrule
        Linear & 95.4\% & 94.4\% & .682 & .377 & 98.5\% & 98.4\% & .618 & .359 & 86.4\% & 96.0\% \\
        Cosine* & \textbf{95.9\%} & \textbf{95.0\%} & \textbf{.675} & \textbf{.370} & \textbf{98.5\%} & \textbf{98.7\%} & \textbf{.600} & \textbf{.347} & \textbf{86.4\%} & \textbf{96.4\%}\\
        \bottomrule
    \end{tabular}
    }
    \label{tab:interpolation}
    % \vspace{-4mm}
\end{table*}

\begin{table*}[t]
    \centering
    \setlength{\tabcolsep}{5pt}
    \caption{Ablation experiments on the effect of different sampling steps.}
    \vspace{-2mm}
    \resizebox{0.9\textwidth}{!}{
    \begin{tabular}{lcccccccccc}
        \toprule
        \multirow{2}[1]{*}{Step} & \multicolumn{4}{c}{Decoding} & \multicolumn{4}{c}{Encoding ($\rightarrow$Decoding)} & \multicolumn{2}{c}{Retrieval}\\
        \cmidrule(lr){2-5} \cmidrule(l){6-9} \cmidrule(l){10-11}
         & Incep$\uparrow$ & CLIP$\uparrow$ & Eff$\downarrow$ & SwAV$\downarrow$ & Incep$\uparrow$ & CLIP$\uparrow$ &Eff$\downarrow$ & SwAV$\downarrow$ & Raw$\uparrow$ & Syn$\uparrow$ \\
        \midrule
        10 & 95.0\% & 94.3\% & .682 & .385 & 98.0\% & 98.3\% & .617 & .356 & 86.4\% & 96.0\% \\
        20* & \textbf{95.9\%} & 95.0\% & .675 & \textbf{.370} & \textbf{98.5\%} & \textbf{98.7\%} & .600 & .347 & \textbf{86.4\%} & \textbf{96.4\%}\\
        30 & 95.7\% & \textbf{95.3\%} & \textbf{.674} & \textbf{.372} & \textbf{98.5\%} & \textbf{98.7\%} & \textbf{.597} & \textbf{.345} & \textbf{86.4\%} & \textbf{96.4\%}\\
        \bottomrule
    \end{tabular}
    }
    \label{tab:steps}
    \vspace{-4mm}
\end{table*}

\paragraph{Ablation Study on Sampling Step.}
We examined the effect of different sampling steps on NeuroFlow’s performance using the Euler solver. As shown in Tab.~\ref{tab:steps}, the model remains robust across varying step counts, with 20 steps providing an optimal balance between accuracy and computational efficiency. Increasing to 30 steps offers only marginal gains, indicating that 20 steps are sufficient to achieve high-quality results while maintaining efficiency.

\section{Baseline Models}
\label{sec:baseline}

\subsection{Framework}

As illustrated in Sec.~\ref{sec:ablation}, a nontrivial \textbf{\textit{modality gap}} remains between neural and visual distributions after contrastive learning in the first stage (i.e., \textbf{NeuroVAE}, or w/o $\mathcal{L}_{\text{XFM}}$.), leading to distorted image reconstructions.
To evaluate the contribution of the proposed XFM module in bridging this gap, we construct two baseline configurations that replace XFM with simpler mechanisms:

\textbf{i) NeuroVAE + MSE.}
This variant substitutes XFM with a direct mean-squared error loss between the neural and visual latent codes. It represents a naïve regression-based alignment strategy and tests whether pointwise matching alone is sufficient to reduce the modality discrepancy.

\textbf{ii) NeuroVAE + LRs.}
This variant replaces XFM with two \textit{independent} linear projection networks mapping between neural and visual latent spaces. As a non-unified framework, it provides a direct comparison point for evaluating the advantage of our unified XFM formulation in bridging the modality gap and maintaining encoding–decoding consistency.

By contrasting these baselines with \textbf{NeuroFlow (NeuroVAE + XFM)}, which provides a \textit{single, unified flow-based transformation} between neural and visual latent distributions, we can directly quantify the benefits of a shared latent structure for improving encoding–decoding consistency and for effectively bridging the neural–visual modality gap.

\subsection{Results}
\begin{figure*}[!t]
\begin{center}
\centerline{\includegraphics[width=0.93\textwidth]{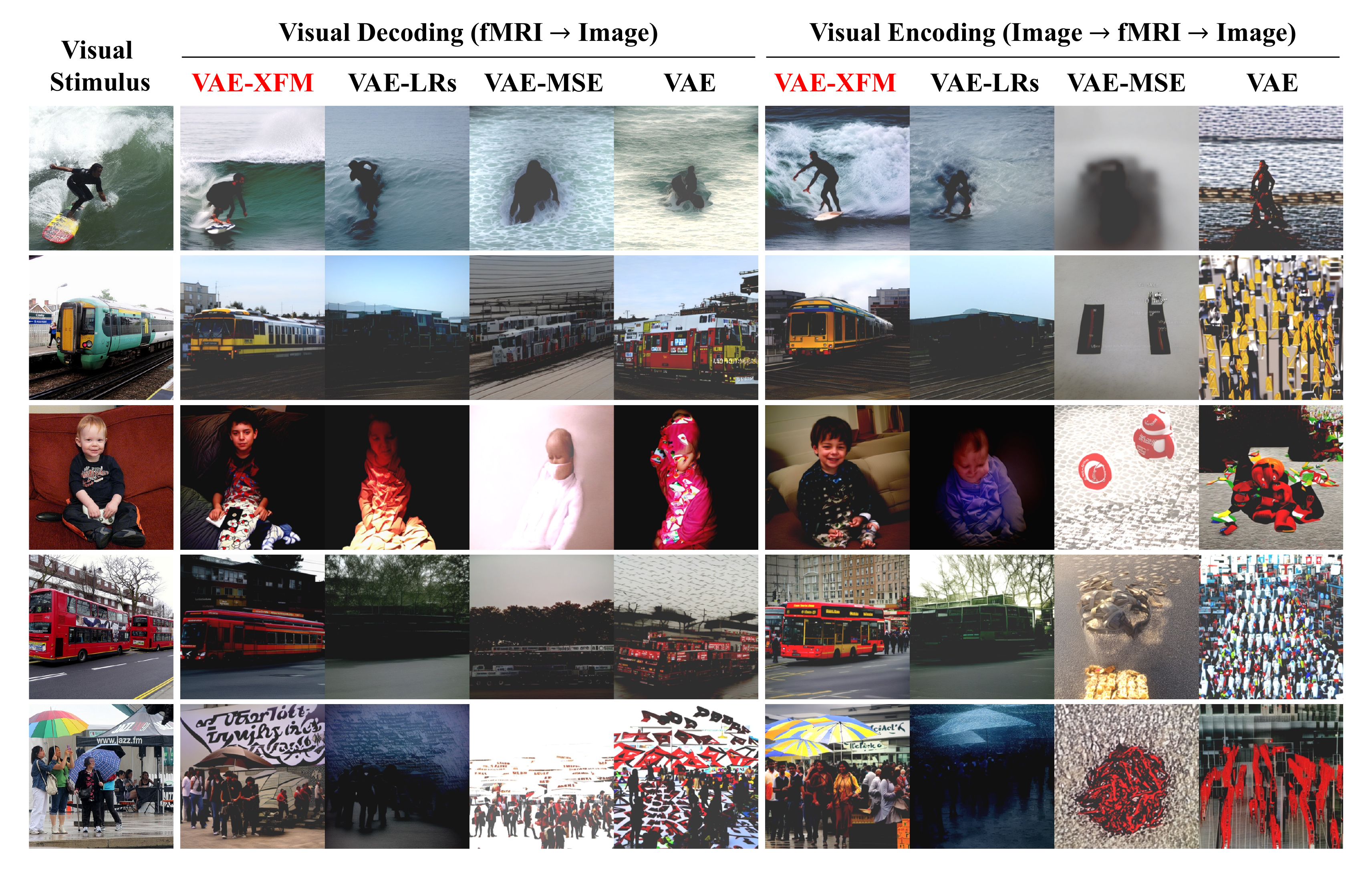}}
\vspace{-3mm}
\caption{Qualitative comparisons between NeuroFlow (NeuroVAE-XFM) and baseline models. 
}
\label{fig:baseline_sup}
\end{center}
% \vskip -0.3in
\end{figure*}

We present the quantitative and qualitative comparisons in Tab.~\ref{tab:baseline} and Fig.~\ref{fig:baseline_sup}. The results reveal clear differences between the baseline configurations and demonstrate the advantages of the proposed cross‐modal flow matching (XFM) framework.

As shown in Tab.~\ref{tab:baseline}, incorporating a simple MSE objective (\textbf{NeuroVAE + MSE}) does not improve performance; instead, it causes noticeable degradation across decoding, encoding, and retrieval metrics. This suggests that pointwise supervision in a shared latent space is inadequate for aligning neural and visual representations and may even exacerbate the existing modality gap.

Replacing XFM with two independent linear projection networks (\textbf{NeuroVAE + LRs}) produces a mixed pattern of results. While decoding performance decreases slightly, both encoding accuracy and retrieval improve substantially. This indicates that linear mappings provide a more flexible alignment mechanism than a pointwise MSE loss. However, \textbf{\textit{a linear network is not sufficient for modeling the complexities inherent in the latent distributions}}, and the independence of the forward and backward transforms prevents them from operating as a unified cross‐modal mapping. As a consequence, the modality gap is only partially reduced, and the resulting encoding–decoding consistency remains limited.

Moving to a unified transformation framework, \textbf{NeuroVAE + XFM} (NeuroFlow) achieves the best performance across all evaluation metrics, including decoding, encoding-decoding consistency, and retrieval. The substantial gains highlight the effectiveness of XFM in \textbf{\textit{bridging the neural–visual modality gap through a coherent flow-based alignment, while simultaneously preserving strong encoding–decoding consistency}}.

These trends are also evident qualitatively in Fig.~\ref{fig:baseline_sup}. NeuroFlow produces reconstructions that better preserve global structure and fine-grained visual semantics in both decoding (fMRI→Image) and encoding (Image→fMRI→Image), whereas \textbf{NeuroVAE}, \textbf{NeuroVAE + MSE} , and \textbf{NeuroVAE + LRs} baselines show \textit{varying degrees of blurriness, structural distortion, or semantic drift}. Together, these results demonstrate that unified flow-matching alignment is essential for achieving high-fidelity cross-modal generation.

\begin{table*}[t]
    \centering
    % \captionsetup{font=small}
    \setlength{\tabcolsep}{4pt}
    \caption{Quantitative comparison between NeuroFlow (NeuroVAE+XFM*) and baseline models.}
    \resizebox{0.9\textwidth}{!}{
    \begin{tabular}{lcccccccccc}
        \toprule
        \multirow{2}[1]{*}{Method} & \multicolumn{4}{c}{Decoding} & \multicolumn{4}{c}{Encoding$\rightarrow$Decoding} & \multicolumn{2}{c}{Retrieval}\\
        \cmidrule(lr){2-5} \cmidrule(l){6-9} \cmidrule(l){10-11}
         & Incep$\uparrow$ & CLIP$\uparrow$ & Eff$\downarrow$ & SwAV$\downarrow$ & Incep$\uparrow$ & CLIP$\uparrow$ &Eff$\downarrow$ & SwAV$\downarrow$ & Raw$\uparrow$ & Syn$\uparrow$ \\
        \midrule
        NeuroVAE & 87.7\% & 83.7\% & .788 & .497 & 60.3\% & 58.1\% & .953 & .644 & 86.4\% & 14.1\% \\
        NeuroVAE+MSE & 85.2\% & 81.4\% & .832 & .536 & 51.0\% & 52.0\% & .978 & .671 & 86.4\% & 2.8\%\\
        NeuroVAE+LRs & 86.8\% & 83.0\% & .794 & .496 & 90.3\% & 87.0\% & .783 & .493 & 86.4\% & 90.8\%\\
        \midrule
        NeuroVAE+XFM* & \textbf{95.9\%} & \textbf{95.0\%} & \textbf{.675} & \textbf{.370} & \textbf{98.5\%} & \textbf{98.7\%} & \textbf{.600} & \textbf{.347} & \textbf{86.4\%} & \textbf{96.4\%}\\
        \bottomrule
    \end{tabular}
    }
    \label{tab:baseline}
    \vspace{-4mm}
\end{table*}

\section{Additional Results}

\subsection{Subject-specific Results}

Tab.~\ref{tab:specific} presents quantitative visual encoding and decoding results for four subjects (Sub1, Sub2, Sub5, Sub7). NeuroFlow consistently achieves high performance across decoding metrics (Inception, CLIP, Eff, SwAV), encoding-decoding consistency, and retrieval accuracy, demonstrating robust performance on different subjects. While absolute scores vary due to individual neural differences, overall trends remain stable, highlighting reliable model performance.

\begin{figure*}[!t]
\begin{center}
\centerline{\includegraphics[width=0.85\textwidth]{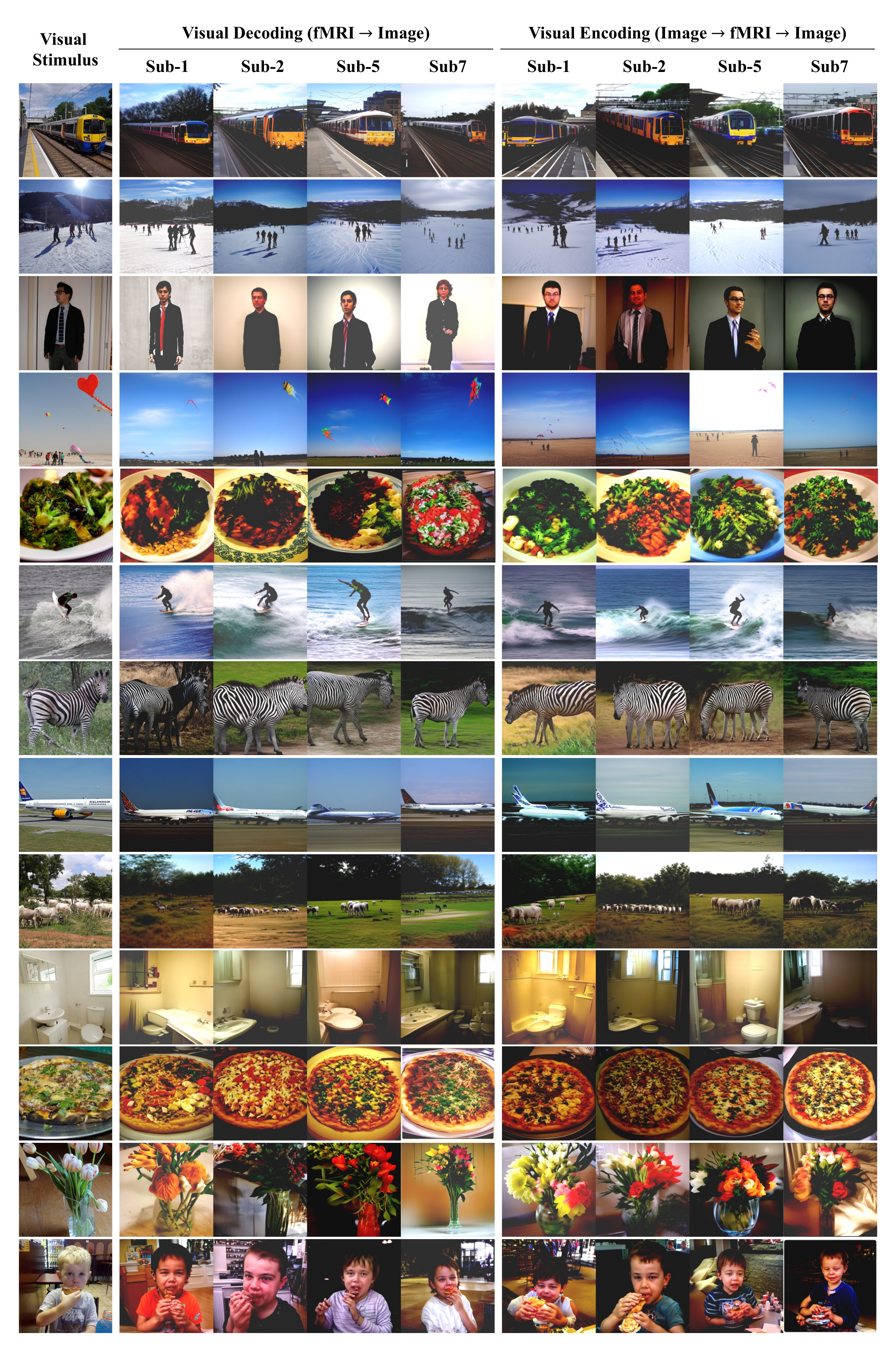}}
\vspace{-1mm}
\caption{Qualitative subject-specific visual encoding and decoding results. 
}
\label{fig:result_sup}
\end{center}
% \vskip -0.3in
\end{figure*}

Qualitative results in Fig.~\ref{fig:result_sup} show visual reconstructions for Sub1, Sub2, Sub5, and Sub7, including direct decoding from fMRI and encoding-decoding cycles (image $\rightarrow$ fMRI $\rightarrow$ image). These examples confirm that NeuroFlow preserves semantic content and fine-grained visual details across subjects, maintaining strong encoding-decoding consistency and further demonstrating the robustness and generalizability of the model.

\begin{table*}[t]
    \centering
    % \captionsetup{font=small}
    \setlength{\tabcolsep}{4pt}
    \caption{Quantitative subject-specific visual encoding and decoding results.}
    \resizebox{0.9\textwidth}{!}{
    \begin{tabular}{lcccccccccc}
        \toprule
        \multirow{2}[1]{*}{Method} & \multicolumn{4}{c}{Decoding} & \multicolumn{4}{c}{Encoding$\rightarrow$Decoding} & \multicolumn{2}{c}{Retrieval}\\
        \cmidrule(lr){2-5} \cmidrule(l){6-9} \cmidrule(l){10-11}
         & Incep$\uparrow$ & CLIP$\uparrow$ & Eff$\downarrow$ & SwAV$\downarrow$ & Incep$\uparrow$ & CLIP$\uparrow$ &Eff$\downarrow$ & SwAV$\downarrow$ & Raw$\uparrow$ & Syn$\uparrow$ \\
        \midrule
        Sub1 & 95.9\% & 95.0\% & .675 & .370 & 98.5\% & 98.7\% & .600 & .347 & 86.4\% & 96.4\%\\
        Sub2 & 95.4\% & 93.3\% & .677 & .362 & 98.5\% & 98.5\% & .595 & .344 & 81.3\% & 97.8\%\\
        Sub5 & 97.0\% & 95.9\% & .645 & .345 & 98.8\% & 99.0\% & .569 & .329 & 84.7\% & 98.3\%\\
        Sub7 & 93.9\% & 92.7\% & .699 & .378 & 98.5\% & 98.5\% & .598 & .344 & 69.8\% & 95.5\%\\
        \bottomrule
    \end{tabular}
    }
    \label{tab:specific}
    % \vspace{-4mm}
\end{table*}

\begin{figure*}[!t]
\begin{center}
\centerline{\includegraphics[width=0.93\textwidth]{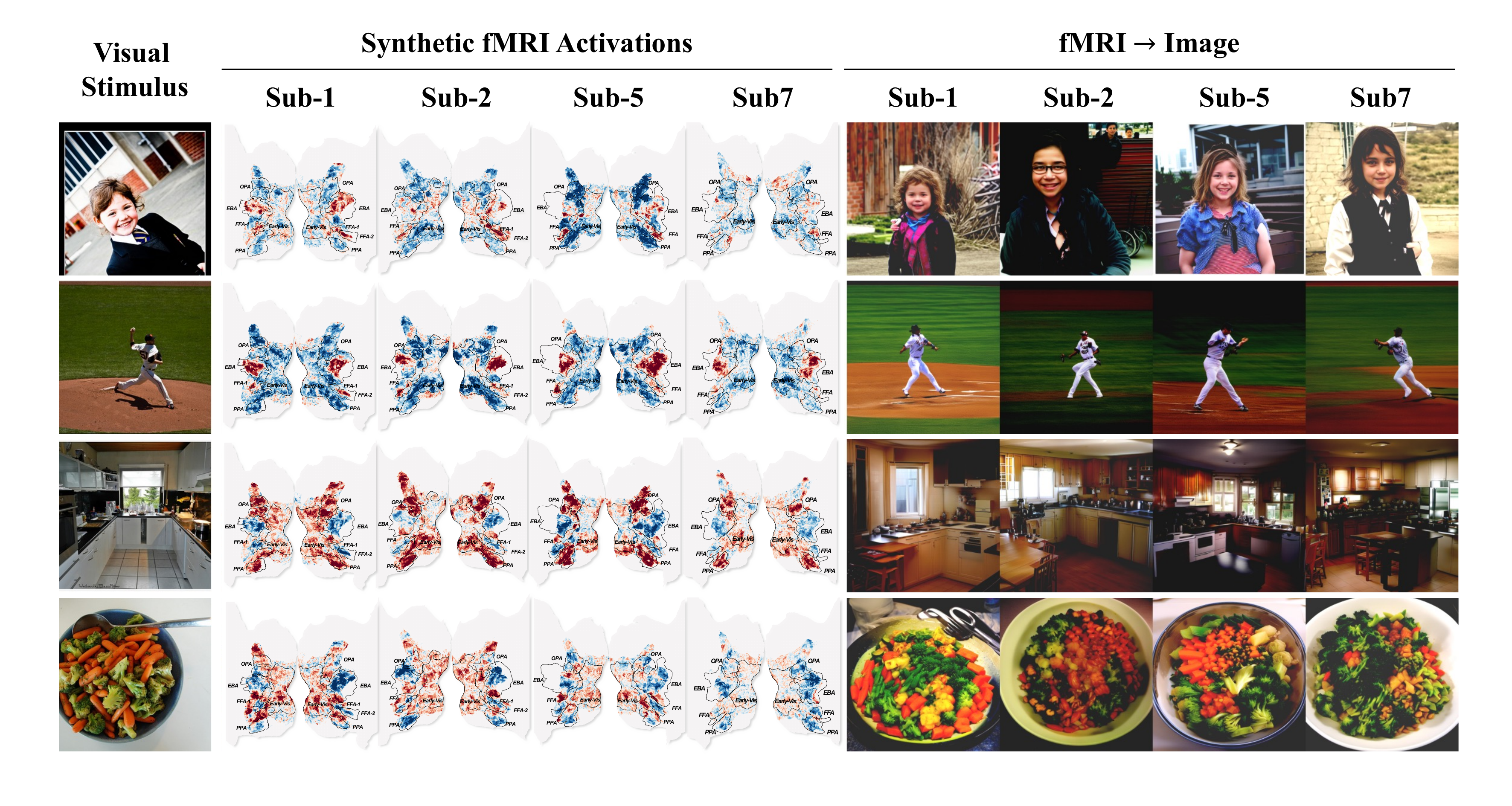}}
\vspace{-1mm}
\caption{Subject-specific fMRI activations and corresponding image reconstructions. 
}
\label{fig:brain_sup}
\end{center}
% \vskip -0.3in
\end{figure*}

\begin{figure*}[!t]
\begin{center}
\centerline{\includegraphics[width=0.93\textwidth]{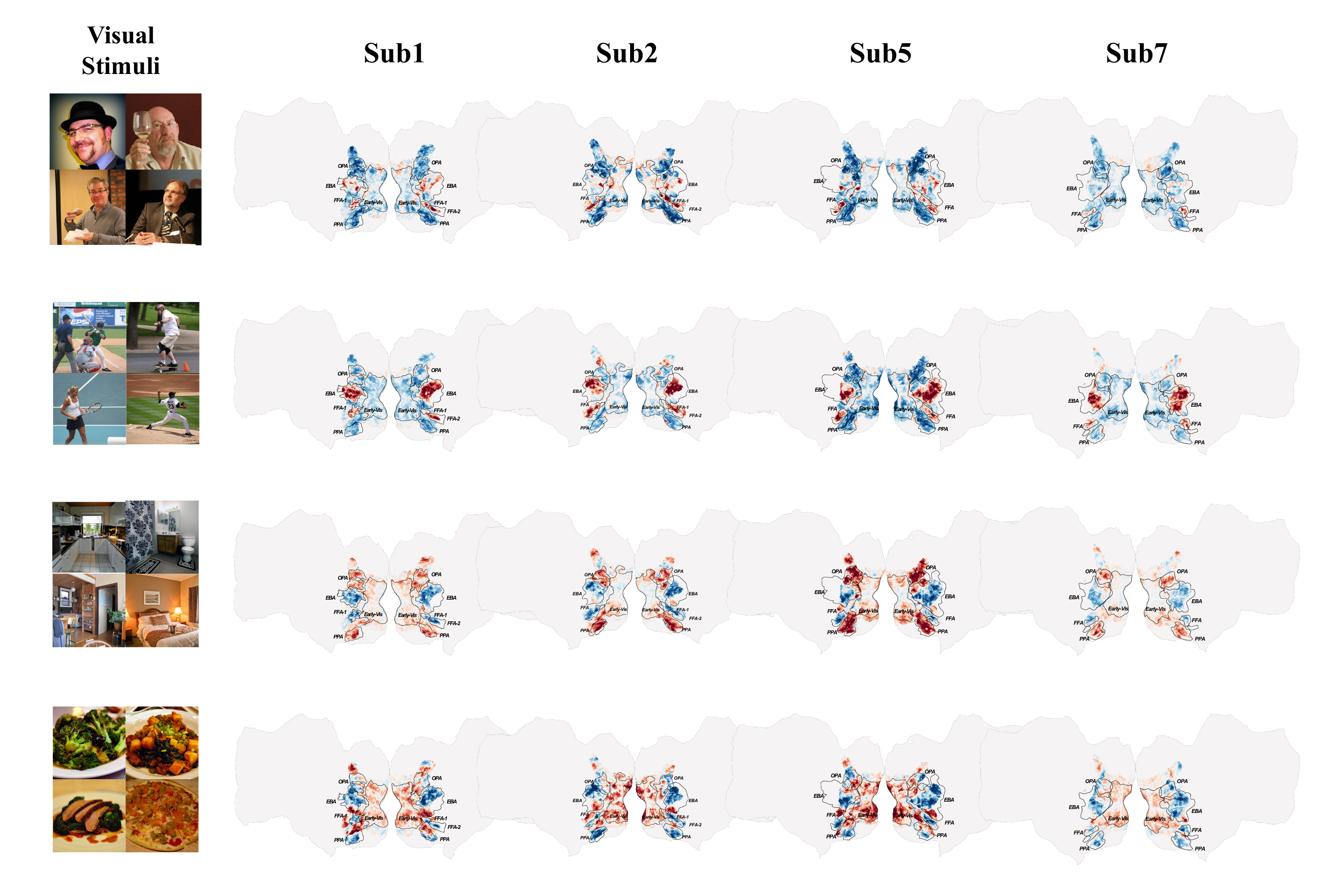}}
\vspace{-1mm}
\caption{Subject-specific category-selective fMRI activations: Faces, Bodies, Places, and Food. 
}
\label{fig:category-sup}
\end{center}
% \vskip -0.3in
\end{figure*}

\subsection{Brain Functional Analysis}

As illustrated in Fig.~\ref{fig:brain_sup}, we present subject-specific fMRI activations generated by NeuroFlow along with their corresponding fMRI-to-image reconstructions across different functional domains, including face, body, place, and food-related regions. Despite substantial inter-subject variability in the spatial distribution of brain activations, NeuroFlow consistently focuses on the appropriate functional areas, such as the fusiform face area (FFA) for face stimuli, while generating semantically coherent reconstructions.

Furthermore, Fig.~\ref{fig:category-sup} shows subject-specific category-selective fMRI activations for Faces, Bodies, Places, and Food. These results reinforce that NeuroFlow reliably captures functional specificity across individuals, attending to the corresponding brain regions for each stimulus category and preserving consistent semantic content in the generated images.

\end{document}